%% file: main.tex
\ifdefined\XeTeXversion\else\ifdefined\directlua\else\pdfoutput=1\fi\fi
\documentclass[11pt]{article}
\usepackage[numbers]{thefinai}
\usepackage{pifont,gradient-text,placeins,float}

\newcommand{\anonlink}[1]{anonymized resource}

\colorlet{CmpGreen}{green!60!black}
\colorlet{CmpRed}{red}
\definecolor{CmpOrange}{HTML}{BB842D}
\definecolor{CmpHeader}{HTML}{EAF0F7}
\definecolor{CmpHighlight}{HTML}{E7F2FA}
\newcommand{\cmpY}{\textcolor{CmpGreen}{\ding{51}}}
\newcommand{\cmpN}{\textcolor{CmpRed}{\ding{55}}}
\newcommand{\cmpP}{\textcolor{CmpRed}{\textbullet}}
\newcommand{\cmpHead}[1]{\textbf{#1}}
\newcommand{\bench}{\textsc{\gradientRGB{FinDendro}{0,0,185}{240,0,15}}\xspace}
\providecommand{\Description}[2][]{ }

\title{LLMs Don't Know What They Know: Benchmarking Temporal Self-Knowledge for Financial Backtesting}
\author{Lingfei~Qian\textsuperscript{1}, Yan~Wang\textsuperscript{1}, Xueqing~Peng\textsuperscript{1}, Yupeng~Cao\textsuperscript{1}, Yuechen~Jiang\textsuperscript{2}, Zhuohan~Xie\textsuperscript{3}, Anke~Xu\textsuperscript{1}, Wenbo~Cao\textsuperscript{1}, Jiarui~Chen\textsuperscript{1}, Zehan~Li\textsuperscript{4}, Jimin~Huang\textsuperscript{1,2}, Alejandro~Lopez-Lira\textsuperscript{5}, Yuehua~Tang\textsuperscript{5}, Sophia~Ananiadou\textsuperscript{2}, Jian-Yun~Nie\textsuperscript{6}}
\affiliation[1]{The Fin AI}
\affiliation[2]{University of Manchester}
\affiliation[3]{Mohamed bin Zayed University of Artificial Intelligence}
\affiliation[4]{Deepkin AI}
\affiliation[5]{University of Florida}
\affiliation[6]{University of Montreal}
\date{October 30, 2026}

\begin{document}
\begin{abstract}
\input{sections/00_abstract}
\end{abstract}
\maketitle

\input{sections/01_introduction}

\input{sections/02_related_work}

\input{sections/03_benchmark_design}
\input{sections/05_experiments}
\FloatBarrier
\input{sections/08_conclusion}
\input{sections/07_analysis}
\FloatBarrier

\section*{Acknowledgments}
The authors acknowledge The Fin AI community for its research support, feedback, and collaborative environment that contributed to this work. This research was supported by the NVIDIA Academic Grant Program using 32K A100 GPU-hours on Brev.

\bibliographystyle{plainnat}
\bibliography{references}

\appendix
\input{sections/A_data_construction}
\input{sections/B_cutoff_summary}
\input{sections/C_manual_validation}

\input{sections/D_model_details}
\end{document}

%% file: sections/00_abstract.tex
Financial backtests of large language models (LLMs) are meaningful only if the model did not already know the outcomes it is asked to predict. Practitioners guard against this by trusting a vendor's stated knowledge cutoff, by asking the model what it knows, or by instructing it to answer as of a past date. To verify whether any of these safeguards can be trusted, we introduce \bench{}, a benchmark of 2,226 dated U.S. facts from 1999 to 2026, covering monthly macroeconomic series, month-end stock prices and Federal Reserve decisions. Each fact yields four prompts: a value question, \textsc{Recall} and \textsc{Predict} questions that pair each candidate value with a claim of knowing or guessing without a declared date, and a \textsc{Predict} question with a historical declared date. Controls for persistence and dominant outcomes keep chance accuracy near 0.27. Results show that for most frontier models, self-reports understate knowledge. GPT-5 answers 76\% of controlled facts correctly but claims to know 38\%, and answers 61\% of the facts it says it does not know. Knowledge claims can persist beyond the measured knowledge boundary, by up to seven quarters, and a declared date removes most of the claims of prior knowledge while leaving accuracy unchanged. 
Backtests inside a model's training window therefore cannot be certified leak-free by self-report, instruction or stated cutoff. 
These findings motivate behavioral measurement of temporal knowledge boundaries, and \bench{} provides a benchmark and procedure for selecting backtest periods based on demonstrated factual knowledge. 

%% file: sections/01_introduction.tex
\section{Introduction}


Large language models (LLMs) are increasingly used and evaluated for financial forecasting, including stock-price prediction~\cite{lopezlira2023stock}, corporate expectations~\cite{jha2024corporate}, and broader financial applications~\cite{xie2024finben,peng2026multifinben,peng2026herculean}. Their predictive capabilities are commonly assessed through historical backtesting, where models generate forecasts for past periods and their predictions are compared with realized outcomes. Financial backtests benefit from long evaluation periods because market conditions change, economic outcomes are noisy, and short evaluation windows after a recent model's training cutoff may not reliably distinguish predictive skill from chance~\cite{gao2026detecting}. However, historical backtesting introduces a fundamental challenge for LLMs: realized outcomes may already be represented in their training data. When asked to predict a past stock price, inflation rate, or monetary policy decision, a model may reproduce the observed outcome rather than infer it from information available at the forecast date~\cite{lopezlira2025memorization,crane2025totalrecall}. Such hindsight leakage can inflate apparent forecasting performance, undermining the reliability of model comparisons and selection for financial applications.

\input{figures/motivation}

Practice relies on three safeguards against this, which are evaluating only after the vendor-stated cutoff~\cite{chen2025stockbench,forecastdojo2026}, instructing the model to answer as of the forecast origin~\cite{halawi2024forecasting,lopezlira2025memorization}, and querying the model about its self-reported knowledge cutoff~\cite{paleka2025pitfalls,chandak2025scaling}. However, most of the existing works ignore to evaluate the reliability of these safeguards. \emph{Input restriction} supplies only information available at the forecast origin, through dated corpora or time-filtered retrieval, or uses instructions to restrict the model ~\cite{zou2022autocast,halawi2024forecasting,ma2026oracleproto}, but it cannot remove outcomes stored in the parameters, which models recall even when told not to~\cite{lopezlira2025memorization}. \emph{Period selection} evaluates after a stated cutoff or release date~\cite{chen2025stockbench,forecastdojo2026}, but a stated cutoff describes the training data rather than what the model retains. \emph{Boundary detection} locates where knowledge ends from perplexity or behavior~\cite{cheng2024dated,pezik2025llmlagbench,jia2026hindsightbench} and finds effective cutoffs that differ from reported ones, but these works do not ask whether the model itself knows where that end lies accurately, which is precisely what self-reports and declared dates rely on. As shown in Figure~\ref{fig:motivation}, the self-reported and the actual boundary of models might be different.
Some works adopt \emph{Leakage-free designs} to sidestep the question. Yet, prospective evaluation~\cite{karger2025forecastbench,zeng2025futurex,yang2025prophetarena,qian2025ama} must wait for outcomes and cannot evaluate history, and point-in-time models~\cite{drinkall2024timemachine,he2025chronologically,yan2026datedgpt,benhenda2026lookahead} remain small. Therefore, what is missing is a measurement of a model's \textbf{actual cutoff} and its \textbf{model-identified cutoff} on the same facts, against which each safeguard can be tested.

To this end, we introduce \bench{}, a benchmark of which the evaluation targets these three safeguards: vendor-stated cutoffs, model self-reports, and input instructions.  It compares models' factual answers with their claims about what they know. It contains 2,226 dated U.S. facts from 1999 to 2026, covering twelve monthly FRED macroeconomic series, month-end stock prices for 50 large companies, and Federal Reserve rate decisions. For each fact, we construct three tasks. A \textbf{value question} (Task~1) asks the model to select the correct answer. Its performance across periods provides an estimate of the {actual cutoff}, which we compare with the vendor-stated cutoff. A separate \textbf{value and claim question} (Task~2) pairs each candidate answer with either a statement of knowing or an admission of guessing. This allows us to examine whether the model's knowledge claims agree with its factual performance and to estimate the {model-identified cutoff}. Task~2 is asked in \textsc{Recall} and \textsc{Predict} forms without a declared date. A \textbf{dated prediction} (Task~3) uses the Task~2 \textsc{Predict} form with an instruction to answer as of a specified historical date. Comparing Task~3 with Task~2 \textsc{Predict} tests whether the date instruction helps separate known outcomes from genuine predictions. We control for simple strategies such as repeating the previous month's value and for imbalanced policy outcomes. We evaluate four frontier models, two smaller open models, and three families of point-in-time models.

The results reveal limitations in each safeguard. \textbf{First, stated cutoff dates do not necessarily locate the end of a model's effective knowledge.} Llama-3.3-70B's factual accuracy falls to chance after mid-2021, well before its vendor-stated cutoff of December 2023, while its knowledge claims continue until early 2023. \textbf{Second, models' self-reports do not reliably distinguish known from unknown outcomes.} GPT-5 answers 76\% of controlled facts correctly but claims to know only 38\%. Within its measured knowledge window, it still answers 75\% of the facts it says it does not know correctly, compared with 33\% after that window. Keeping only facts that the model claims not to know therefore does not reliably remove leakage. \textbf{Third, historical-date instructions can change knowledge claims without materially changing factual accuracy.} Asking GPT-5 to answer as of a past date reduces its claims of prior knowledge from 39\% to 1\%, while accuracy remains almost unchanged. Outcomes the model can answer may consequently be presented as predictions despite these instructions. The small point-in-time models offer a different limitation: their accuracy remains near chance on either side of their training cutoffs. Together, these experiments show why backtesting safeguards should be checked against measured model behavior. \bench{} contributes a benchmark that pairs factual answers with knowledge claims, a consistent procedure for estimating both temporal boundaries, and direct tests of the safeguards used in historical backtesting.

%% file: figures/motivation.tex
\begin{figure}[t]
\centering
\includegraphics[width=\columnwidth]{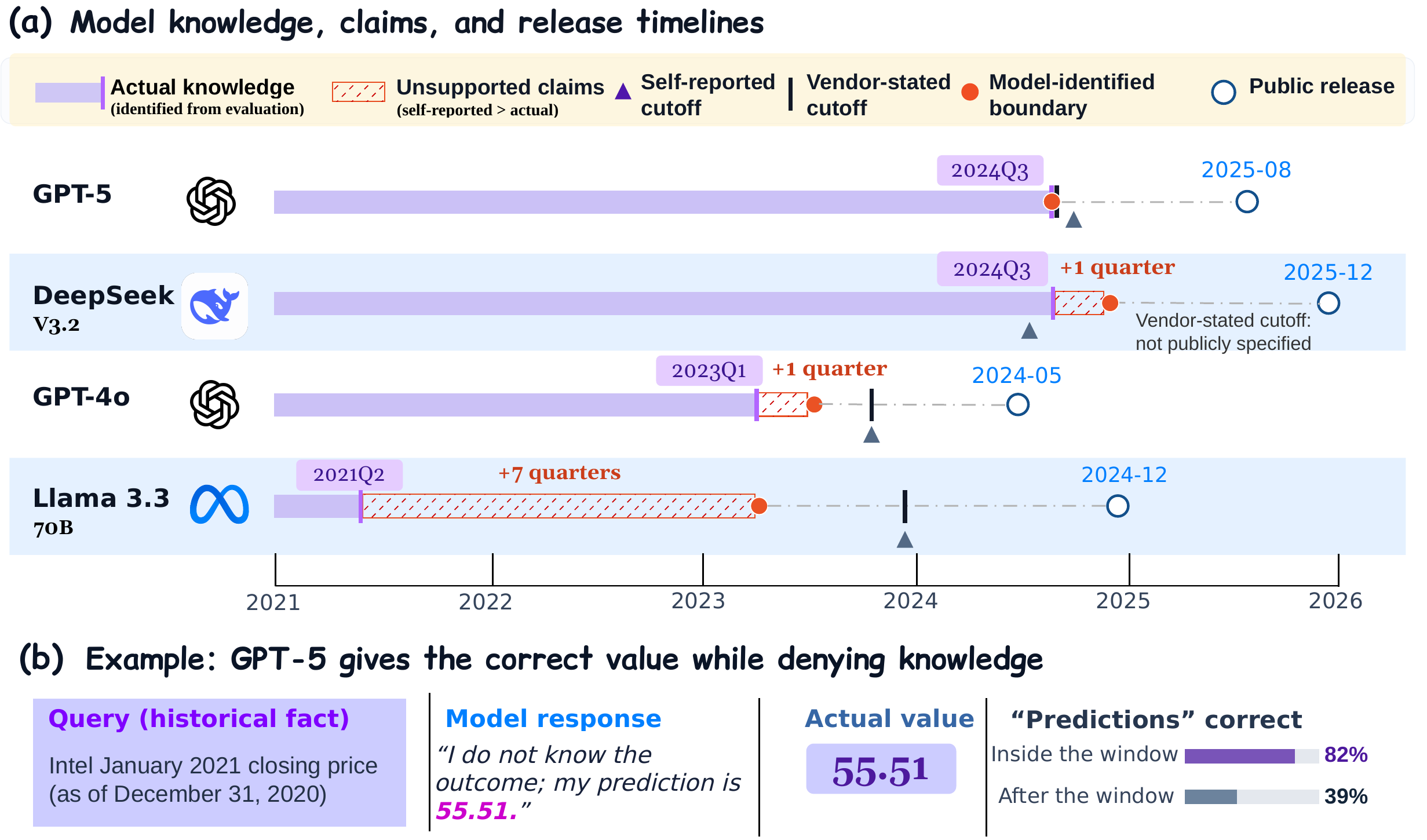}
\caption{\textbf{Model knowledge, claims, and release timelines.} 
(a) Knowledge and claim boundaries measured by \bench{}, compared with self-reported and vendor-stated cutoffs and public release dates. 
(b) A historical example in which GPT-5 denies knowing an outcome but produces the correct value, illustrating why self-reported knowledge boundaries cannot reliably prevent leakage in retrospective forecasting.}
\label{fig:motivation}
\Description{Four horizontal timelines from 2021 to 2026, one per model. GPT-5's knowledge and claims both end in the third quarter of 2024, matching its vendor-stated cutoff, and it was released in August 2025. DeepSeek-V3.2's knowledge ends in the third quarter of 2024 and its claims one quarter later, and it states no cutoff. GPT-4o's knowledge ends in early 2023, claims one quarter later, and its stated cutoff is October 2023. Llama-3.3-70B's knowledge ends in mid-2021 while its claims continue for seven more quarters, close to its stated cutoff of December 2023. Below, a box shows GPT-5, told that today is December 31, 2020, saying it does not know Intel's January 2021 closing price and predicting 55.51, the actual value, and such predictions are right 82 percent of the time inside its knowledge window and 39 percent after it.}
\end{figure}

%% file: sections/02_related_work.tex
\section{Related Work}

\begin{table}[t]
\centering
\small
\setlength{\tabcolsep}{2.8pt}
\renewcommand{\arraystretch}{1.16}

\caption{Comparison with related benchmarks.
\emph{Fin.} denotes financial or economic content;
\emph{Dated}, reference-dated facts;
\emph{Cutoff}, behavioral knowledge-boundary estimation;
\emph{Claims}, explicit knowledge claims;
and \emph{Guards}, evaluation of leakage safeguards.
\cmpY{} = yes, \cmpP{} = partial, and
\cmpN{} = no.}

\label{tab:compare}

\begin{tabular}{@{}lccccc@{}}
\toprule
\rowcolor{CmpHeader}
\cmpHead{Benchmark} &
\cmpHead{Fin.} &
\cmpHead{Dated} &
\cmpHead{Cutoff} &
\cmpHead{Claims} &
\cmpHead{Guards} \\
\midrule

FinBen~\cite{xie2024finben}
& \cmpY & \cmpN & \cmpN & \cmpN & \cmpN \\

StockBench~\cite{chen2025stockbench}
& \cmpY & \cmpN & \cmpN & \cmpN & \cmpN \\

FinTMMBench~\cite{zhu2025fintmmbench}
& \cmpY & \cmpY & \cmpN & \cmpN & \cmpN \\

ForecastBench~\cite{karger2025forecastbench}
& \cmpP & \cmpN & \cmpN & \cmpN & \cmpN \\

Look-Ahead-Bench~\cite{benhenda2026lookahead}
& \cmpY & \cmpP & \cmpN & \cmpN & \cmpP \\

SelfAware~\cite{yin2023selfaware}
& \cmpN & \cmpN & \cmpN & \cmpY & \cmpN \\

EvolveBench~\cite{zhu2025evolvebench}
& \cmpN & \cmpY & \cmpN & \cmpP & \cmpN \\

Dated Data~\cite{cheng2024dated}
& \cmpN & \cmpY & \cmpY & \cmpN & \cmpN \\

LLMLagBench~\cite{pezik2025llmlagbench}
& \cmpN & \cmpY & \cmpY & \cmpN & \cmpN \\

HindsightBench~\cite{jia2026hindsightbench}
& \cmpY & \cmpY & \cmpY & \cmpN & \cmpP \\

\midrule
\rowcolor{CmpHighlight}
\textbf{\bench{} (ours)}
& \cmpY & \cmpY & \cmpY & \cmpY & \cmpY \\

\bottomrule
\end{tabular}
\end{table}

\subsection{LLM Forecasting and Financial Evaluation}

Existing LLM forecasting benchmarks address temporal validity through information restrictions or evaluation-period selection. Autocast uses dated news corpora to restrict information available at prediction time~\cite{zou2022autocast}, while AutoCast++ improves evidence retrieval~\cite{yan2024autocastpp}. Halawi et al.~\cite{halawi2024forecasting} combine retrieval, reasoning, and aggregation, and OracleProto reconstructs historical forecasting conditions through temporally masked retrieval~\cite{ma2026oracleproto}. Prospective benchmarks such as ForecastBench evaluate predictions before outcomes are known~\cite{karger2025forecastbench}. In finance, FinBen, MultiFinBen, and Herculean cover broad financial tasks and agent workflows~\cite{xie2024finben,peng2026multifinben,peng2026herculean}. More specialized benchmarks for taxonomy-grounded financial information extraction~\cite{wang2025fintagging} and multi-document financial auditing~\cite{wang2026finauditing}. StockBench evaluates trading beyond reported knowledge cutoffs~\cite{chen2025stockbench}, while FinTMMBench focuses on temporal financial retrieval~\cite{zhu2025fintmmbench}. Collectively, these benchmarks assess financial capabilities or forecasting performance, but do not directly evaluate whether models can reliably identify historical outcomes they already know.

\subsection{Look-ahead Bias in LLM Backtests}

Prior work has examined look-ahead bias in financial LLM evaluations from several perspectives. Glasserman and Lin~\cite{glasserman2023lookahead} use anonymized headlines to separate hindsight effects from company knowledge. Lopez-Lira et al.~\cite{lopezlira2025memorization} demonstrate the reproduction of historical economic and market outcomes, while Crane et al.~\cite{crane2025totalrecall} quantify macroeconomic recall. Gao et al.~\cite{gao2026detecting} propose tests for look-ahead bias, and Paleka et al.~\cite{paleka2025pitfalls} examine broader pitfalls in evaluating LLM forecasters. Other approaches mitigate leakage through point-in-time training~\cite{drinkall2024timemachine,he2025chronologically,yan2026datedgpt}, while Look-Ahead-Bench compares trading performance across in-sample and out-of-sample periods~\cite{benhenda2026lookahead}. These studies establish the importance of hindsight leakage but do not jointly examine factual recall and models' explicit knowledge claims.

\subsection{Knowledge Boundaries and Self-Knowledge}

Research on model self-knowledge examines whether LLMs can recognize their own knowledge limitations. Kadavath et al.~\cite{kadavath2022know} investigate models' ability to assess answer correctness, while SelfAware evaluates recognition of unanswerable questions~\cite{yin2023selfaware}. Temporal benchmarks such as TempLAMA, DyKnow, and EvolveBench assess evolving factual knowledge~\cite{dhingra2022timeaware,mousavi2024dyknow,zhu2025evolvebench}, while ChronoRegime examines whether historical evidence supports future predictions~\cite{huang2026predictable}. Complementary work estimates temporal knowledge boundaries: Dated Data uses perplexity over dated documents~\cite{cheng2024dated}, LLMLagBench probes recent-event knowledge~\cite{pezik2025llmlagbench}, and HindsightBench estimates behavioral cutoffs through time-indexed macroeconomic questions and date manipulation~\cite{jia2026hindsightbench}. These studies investigate knowledge awareness or temporal boundaries, but do not jointly compare explicit knowledge claims with factual performance across the same dated financial facts.

Table~\ref{tab:compare} summarizes the differences between \bench{} and existing benchmarks. While prior work addresses financial evaluation, temporal knowledge, or self-awareness, \bench{} combines dated financial facts, behavioral knowledge-boundary estimation, and fact-level knowledge claims to examine the reliability of safeguards used in historical backtesting.

%% file: sections/03_benchmark_design.tex
\section{The \bench{} Benchmark}
\label{sec:benchmark}

\bench{} asks two questions of any model. Up to which date does it know dated financial facts, and up to which date does it say that it does? Three design choices follow from this aim. First, all facts come from one economy and one calendar, so that a change in behavior over time reflects the model rather than a change in the material. Second, every fact is scored against the value published at the time, so that a correct answer is evidence of recall rather than of a later revision. Third, the same candidate values are offered with and without a statement of knowledge, so that a model's answer and its claim about that answer come from a single choice.

\subsection{Fact collection}
\label{sec:facts}

The pool contains 20,972 dated U.S. facts from January 1999
to September 2026 in three families. Macroeconomic facts
are monthly values of twelve FRED series, namely headline
and core CPI inflation, unemployment, the effective federal
funds rate, the 3-month, 2-year and 10-year Treasury rates,
the 30-year mortgage rate, three dollar exchange rates,
and WTI oil. We use first-release observations retrieved, retaining the initial publication date
and value rather than subsequent revisions.\footnote{
\url{https://alfred.stlouisfed.org/}.}
Company facts are month-end closing prices of 50 large
U.S. companies obtained from Yahoo Finance, using the last trading day of each
month and reversing subsequent stock-split adjustments
to recover prices as quoted at the time.\footnote{
\url{https://finance.yahoo.com/};
\url{https://ranaroussi.github.io/yfinance/}.}
Policy facts are monthly Federal Reserve target-rate outcomes, derived from the
target-rate history available through FRED.\footnote{
\url{https://fred.stlouisfed.org/series/DFEDTAR};
\url{https://fred.stlouisfed.org/series/DFEDTARL};
\url{https://fred.stlouisfed.org/series/DFEDTARU}.}
For each fact, we score against the value published at the time.



We collected 2,226 facts in total, of which 1,353 are macroeconomic, 540 company and 333 policy facts. Sampling is sparse before 2021 (two macroeconomic facts and one stock per month, plus every FOMC month), giving 1,056 facts, and dense from 2021 onward (every macroeconomic series and four stocks per month), giving 1,170 facts, because the cutoffs of current models fall between 2021 and 2025. Table~\ref{tab:stats} summarizes the sample. Of the 2,226 facts, 1,340 count toward knowledge after the controls of Section~\ref{sec:controls}. Each fact yields four prompts, for 8,904 prompts per model, and since every answer is a single option letter, evaluating a new model is inexpensive. The full pool, the sample, the question files and the construction code will be released with the benchmark.

\begin{table}[t]
\centering
\small
\setlength{\tabcolsep}{4pt}
\caption{The evaluation sample by fact family. \emph{Controlled} counts the facts that count toward knowledge, which are the numerical facts where repeating last month's value fails and all policy facts.}
\label{tab:stats}
\begin{tabular}{@{}lrrrrr@{}}
\toprule
Family & Series & Facts & Before 2021 & 2021 on & Controlled\\
\midrule
Macroeconomic & 12 & 1,353 & 528 & 825 & 715\\
Stock prices & 50 & 540 & 264 & 276 & 292\\
Fed decisions & 1 & 333 & 264 & 69 & 333\\
\midrule
Total & 63 & 2,226 & 1,056 & 1,170 & 1,340\\
\bottomrule
\end{tabular}
\end{table}

\subsection{Question and Answer Construction}
\label{sec:task-definitions}

Based on the collected facts, we construct three aligned
tasks to evaluate models' temporal self-knowledge.
Task~1 asks the model to identify the correct value,
providing a behavioral measure of factual knowledge.
Task~2 additionally elicits a claim about whether the
model knows that value, using \textsc{Recall} and
\textsc{Predict} wordings without a declared date.
Comparing Task~1 and Task~2 allows
us to assess whether knowledge claims agree with factual
performance and, across reference periods, to estimate
the actual and model-identified cutoffs.
Task~3 adds a historical-date instruction to the
\textsc{Predict} wording to test its effect on answers
and knowledge claims.
As shown in Table~\ref{tab:example}, each fact yields
four prompts that share the same candidate values.
Numerical facts offer four values spaced by roughly
one typical monthly change of the series, with the
rank of the true value drawn uniformly, and policy
facts offer the three outcomes. Every prompt asks the
model to answer from its own knowledge, without browsing
or tools, and to output a single option letter.

\paragraph{Task 1 (value).} We first establish a behavioral reference for what the
model knows about the collected facts. For each fact,
the model selects one of the candidate values without
making a knowledge claim. Its factual accuracy across
reference periods provides the basis for estimating
the actual cutoff.

\paragraph{Task 2 (value and claim).}
We then examine whether the model recognizes the knowledge
measured in Task~1. To this end, Task~2 asks the model to
select a candidate value together with a claim about
whether it knows that value. 
Each candidate appears twice:
once with a claim of knowing and once with a claim of not
knowing, giving eight options for numerical facts and six
for policy facts.
We use two wordings, both without a declared date,
to examine how prediction framing affects knowledge claims.
\textsc{Recall} asks what the value was,
using ``I know this value'' and ``I do not know this value''
followed by a best guess. \textsc{Predict} asks the model
to predict the value, using ``I already know the outcome''
and ``I do not know the outcome'' followed by a prediction;
policy outcomes are expressed in the future tense.

\paragraph{Task 3 (dated prediction).}
We next test whether a historical-date instruction changes
factual answers or knowledge claims.
Task~3 uses the same question, candidate values, and paired
claim options as Task~2 \textsc{Predict}, adding the prefix
``Assume that today is [date]. Answer as you would on that
date.'', where [date] is the last day of the month preceding
the reference period.
Comparing Task~3 with Task~2 \textsc{Predict} measures the
effect of this date instruction.

\begin{table}[t]
\centering
\small
\setlength{\tabcolsep}{3pt}
\caption{An illustrative CPI fact across three tasks and four prompts. Candidate values are shared, and options are summarized. Each prompt is sent independently with shuffled options.}
\label{tab:example}
\begin{tabular}{@{}>{\raggedright\arraybackslash}p{0.98\columnwidth}@{}}
\toprule
\textbf{Fact.} June 2023 U.S. headline CPI inflation: the 12-month change in FRED CPIAUCNS (all items, NSA), rounded to one decimal place. \textbf{Candidate values:} 3.0\%, 3.4\%, 3.2\%, 2.8\%.\\
\midrule
\textbf{Task~1: Value}\newline
``What was U.S. headline CPI inflation in June 2023?''\newline
\emph{Options:} the four candidate values.\\
\midrule
\textbf{Task~2: Value and claim}\\[3pt]
\textsc{Recall}: ``Recall the value from your memory: what was U.S. headline CPI inflation in June 2023? If you do not know it, select `I do not know'. \newline
\emph{Options:} each value appears twice: once as known and once as a guess (8 choices).\\
\arrayrulecolor{black!30}\cmidrule[0.25pt](lr){1-1}\arrayrulecolor{black}
\textsc{Predict}: ``Predict U.S. headline CPI inflation in June 2023.''\newline
\emph{Options:} each value appears twice: once as already known and once as a prediction of an unknown outcome (8 choices).\\
\midrule
\textbf{Task~3: Dated prediction}\newline
``Assume that today is May 31, 2023. Answer as you would on that date. Predict U.S. headline CPI inflation in June 2023.''\newline
\emph{Options:} the same paired options as Task~2 \textsc{Predict} (8 choices).\\
\bottomrule
\end{tabular}
\end{table}

\subsection{Controls}
\label{sec:controls}


To reduce the effects of persistence and class imbalance
on measured factual accuracy, we apply two controls.
These address cases where a model can answer correctly
without recalling the particular historical outcome.
For numerical facts, a persistence strategy selects the
candidate closest to the previous month's value. When
the monthly change is smaller than half an option step,
the true value remains the closest candidate, allowing
this strategy to answer correctly without knowing the
current month's outcome. This occurs for 47\% of numerical
facts. Including these cases would risk attributing
accuracy from persistence to factual knowledge.
We therefore measure numerical knowledge only on the
remaining 1,007 facts, for which this strategy fails.
For policy facts, ``unchanged'' is correct in 78\% of
months. Ordinary accuracy would therefore reward a model
that always selects this outcome, even if it cannot
identify rate increases or decreases. We instead use
balanced accuracy, averaging the proportion of correct
answers within each of the three outcome classes so
that each receives equal weight. Always selecting
``unchanged'' then scores one third rather than 78\%.

Together, these controls reduce the contribution of
persistence and dominant outcomes to the knowledge score.
The combined chance reference is approximately 0.27,
reflecting four-option numerical questions and
three-outcome policy questions.

\subsection{Measures}
\label{sec:measures}

For each model we report five quantities.
\emph{Knows} is Task~1 accuracy on the 1,340 controlled facts.
\emph{Says} is the share choosing a ``know'' option
in Task~2 \textsc{Recall} without a declared date,
aggregated over the full sample of 2,226 facts.
We also report \emph{accuracy when the model says it knows}
and \emph{accuracy when the model says it does not know},
measured using Task~1 on facts grouped by these
\textsc{Recall} claims, and the \emph{within-period gap}
between these accuracies.
The gap is computed within each period (five-year blocks
before 2021 and half-years after), weighted by period size,
with 95\% intervals from 1,000 bootstrap resamples of facts.
Separately, we report the change in claims and accuracy
from Task~2 \textsc{Predict} to Task~3.


Based on these quantities, we estimate the actual and
model-identified cutoffs shown in
Figure~\ref{fig:motivation}.
The \textbf{actual cutoff} is estimated
as the last quarter whose
two-quarter rolling accuracy (\emph{Knows}) stays above
the midpoint between the model's 2015 to 2020 level
and chance.
We estimate the \textbf{model-identified cutoff} as the last quarter whose claim rate
(\emph{Says it knows}) stays above half of its
2015 to 2020 level.
When \emph{claims end} falls after \emph{knowledge ends},
the intervening quarters indicate continued knowledge
claims beyond the measured knowledge boundary.
We also record the \textbf{self-reported cutoff}, which is the answer when the model is asked for its cutoff directly, and the \textbf{vendor-stated cutoff}, which is published by the model developer.

%% file: sections/05_experiments.tex
\section{Experiments and Results}
\label{sec:experiments}

\subsection{Setup}
\label{sec:setup}

We evaluate four frontier models via API: GPT-5~\cite{openai2025gpt5}, DeepSeek-V3.2~\cite{deepseek2025v32}, GPT-4o~\cite{openai2024gpt4o}, and Llama-3.3-70B-Instruct~\cite{meta2024llama33}. The reported knowledge cutoffs are September 2024 for GPT-5, October 2023 for GPT-4o, and December 2023 for Llama-3.3-70B; DeepSeek-V3.2 has no publicly stated cutoff. We use temperature 0 where supported and a maximum of 1,024 output tokens, increased to 4,096 for GPT-5 to accommodate its reasoning tokens. Each prompt is evaluated once. For comparison, we also evaluate two locally hosted open models, Qwen3-8B~\cite{yang2025qwen3} and Qwen3.5-9B~\cite{qwen2026qwen35}, using greedy decoding with thinking disabled. We further include three families of point-in-time models trained on text available before fixed cutoff dates: TiMaGPT~\cite{drinkall2024timemachine}, ChronoGPT-Instruct~\cite{he2025chronologically}, and DatedGPT~\cite{yan2026datedgpt}. See appendix~\ref{sec:evaluated-models} for details.

\input{figures/tab_results}

\input{figures/fig_honesty_map}

\input{figures/fig_cutoff_timeline}

\subsection{Frontier models know more than they say}
\label{sec:underreport}


Figure~\ref{fig:honesty} compares models' factual
accuracy on controlled facts in Task~1 with their
knowledge claim rates in Task~2 \textsc{Recall}
without a declared date.
GPT-5, DeepSeek-V3.2, and GPT-4o show substantially
higher factual accuracy than knowledge claim rates.
GPT-5 answers 76\% of controlled facts correctly
but claims to know only 38\%, while GPT-4o achieves
50\% accuracy with a claim rate of 20\%
, as shown in Table~\ref{tab:results}.
In contrast, Llama-3.3-70B's claim rate slightly
exceeds its factual accuracy (0.48 versus 0.43).
This tendency to over-claim is more pronounced
in the two Qwen models, which claim to know about
70\% of facts despite accuracies of 0.36--0.37,
close to the chance level of 0.27.

\subsection{Knowledge claims can outlast factual knowledge}
\label{sec:timeline}

Figure~\ref{fig:timeline} tracks factual accuracy and knowledge claims over time, revealing differences between where models' knowledge ends and where they stop claiming it. As shown in Figure~\ref{fig:motivation}, for GPT-5, both boundaries coincide at 2024Q3, matching its vendor-stated cutoff. In contrast, Llama-3.3-70B shows a seven-quarter gap: its factual accuracy falls to chance after 2021Q2, but it continues claiming knowledge until 2023Q1, closer to its reported cutoff of December 2023. GPT-4o's knowledge ends one quarter before its claims, while DeepSeek-V3.2 shows a similar one-quarter gap. The two Qwen models remain near chance across the timeline but continue claiming knowledge until 2024. These patterns show that knowledge claims do not consistently track the knowledge models have. Therefore, we cannot trust the self-reported knowledge of models.

This mismatch matters for financial backtesting, where reported cutoffs are used to select evaluation periods. For Llama-3.3-70B, the vendor-stated cutoff is ten quarters later than its measured knowledge boundary, excluding a substantial period in which factual accuracy has already declined toward chance. More broadly, reported cutoffs and knowledge claims provide different pictures of the temporal extent of models' factual knowledge.

We also examine the sensitivity of boundary estimates
to the detection thresholds.
Changing the thresholds from one-half to one-third
of the distance between the 2015--2020 baseline and
chance shifts GPT-5's and DeepSeek-V3.2's estimated
boundaries by at most one quarter.
Llama-3.3-70B retains a six-to-seven-quarter gap,
while GPT-4o's estimated gap increases from one
to six quarters.
We additionally test more permissive threshold settings.
These yield unstable boundary estimates as the accuracy
threshold approaches chance, allowing sampling
fluctuations to shift the estimated knowledge boundary.

\input{figures/fig_idk}
\subsection{Self-report does not clean a backtest}
\label{sec:filter}
\label{sec:idk}

We then examine whether knowledge claims distinguish
facts the model answers more accurately from those
it answers less accurately.
Table~\ref{tab:results} compares Task~1 accuracy on facts
the model claims to know and those it claims not
to know in Task~2 \textsc{Recall} without a declared date.
The accuracies are presented
in the \emph{Acc$\mid$says} and
\emph{Acc$\mid$denies} columns, respectively.
For GPT-5, they are 95\% and 61\%.
The \emph{Gap} column reports the difference computed
within each historical period and then weighted
by period size.
Across the four frontier models, these within-period
gaps range from 0.08 for DeepSeek-V3.2 to 0.39 for
GPT-4o, and all 95\% bootstrap confidence intervals
exclude zero.
Knowledge claims therefore carry information about
which facts the model answers more accurately.

However, accuracy on disclaimed facts can still
substantially exceed chance.
GPT-5's accuracy on this group is more than twice
the chance reference of 0.27, and DeepSeek-V3.2
also achieves 48\% accuracy on disclaimed facts.
GPT-4o and Llama-3.3-70B are closer to chance on
disclaimed facts, at 34\% and 29\%, respectively,
compared with 83\% and 58\% on facts claimed as known.
The reliability of ``I don't know'' thus varies
across models.
In particular, an admission of not knowing does
not reliably identify unknown historical outcomes
for the strongest models.

Table~\ref{tab:examples} illustrates how these
admissions appear in evaluations.
In Task~2 \textsc{Recall}, GPT-5 states that it does not
know Intel's January 2021 closing price but selects
the correct value of 55.51.
In Task~3, when instructed to predict the same price as of
December 31, 2020, it again denies prior knowledge
and selects 55.51.
The same pattern appears for Chevron's January
2006 closing price and the February 2024 USD/JPY
exchange rate.
In each case, the correct historical outcome is
presented as a guess or prediction despite the
model's denial of knowledge.
Such answers would count as forecasts in a
historical backtest, with the disclaimer
failing to establish that the outcome was unknown.

\begin{table}[t]
\centering
\small
\setlength{\tabcolsep}{6pt}
\renewcommand{\arraystretch}{1.2}
\caption{Facts GPT-5 says it does not know but answers correctly in Task~2 \textsc{Recall} (without a declared date) and Task~3 (dated prediction, with the date set to the end of the preceding month). Each value is the published one among four candidates.}
\label{tab:examples}

\begin{tabularx}{\linewidth}{@{}>{\raggedright\arraybackslash}p{0.28\linewidth}>{\raggedright\arraybackslash}X>{\raggedright\arraybackslash}X@{}}
\toprule
\textbf{Fact} & \textbf{Task~2: \textsc{Recall}} & \textbf{Task~3: \textsc{Predict}} \\
\midrule
Intel close, Jan.\ 2021 & does not know; guesses \textbf{55.51} & does not know; predicts \textbf{55.51} \\
\addlinespace[3pt]
Chevron close, Jan.\ 2006 & does not know; guesses \textbf{59.38} & does not know; predicts \textbf{59.38} \\
\addlinespace[3pt]
JPY per USD, Feb.\ 2024 avg. & does not know; guesses \textbf{149.62} & does not know; predicts \textbf{149.62} \\
\bottomrule
\end{tabularx}
\end{table}

\begin{table}[t]
\centering
\small
\setlength{\tabcolsep}{4pt}
\caption{Self-report filtering on the 1,340 controlled facts, split at each model's \emph{knowledge ends}. Using Task~2 \textsc{Recall} without a declared date, we report each group's size ($n$), the share disclaimed (\emph{Denied}), and Task~1 accuracy on disclaimed facts (\emph{Acc}; chance $\approx$ 0.27).}
\label{tab:filter}
\begin{tabular}{@{}lrrrrrr@{}}
\toprule
& \multicolumn{3}{c}{\textbf{Inside window}} & \multicolumn{3}{c}{\textbf{After window}}\\
\cmidrule(lr){2-4}\cmidrule(l){5-7}
\textbf{Model} & \textbf{\boldmath$n$} & \textbf{Denied} & \textbf{Acc} & \textbf{\boldmath$n$} & \textbf{Denied} & \textbf{Acc}\\
\midrule
GPT-5           & 1,109 & 0.46 & \textbf{0.75} & 231 & 1.00 & 0.33\\
DeepSeek-V3.2   & 1,109 & 0.57 & {0.53} & 231 & 0.96 & \textbf{0.34}\\
GPT-4o          & 925   & 0.60 & 0.37 & 415 & 0.97 & 0.32\\
Llama-3.3-70B   & 711   & 0.22 & 0.31 & 629 & 0.78 & 0.27\\
\bottomrule
\end{tabular}
\end{table}

We next examine the practical consequences of using
these knowledge claims to select backtest samples.
Specifically, we apply a filter that excludes facts
the model claims to know in \textsc{Recall} and
retains only those it disclaims.
Table~\ref{tab:filter} reports the fraction of facts
retained and their Task~1 accuracy, separately
inside and after each model's measured knowledge
window.
For GPT-5, the filter retains 46\% of in-window
facts, yet accuracy on these retained facts remains
0.75, compared with 0.33 after its knowledge boundary.
DeepSeek-V3.2 shows a similar pattern, with accuracy
of 0.53 on retained in-window facts versus 0.34
afterward.
The corresponding differences are smaller for
GPT-4o and Llama-3.3-70B, whose in-window accuracies
are closer to their post-boundary levels
(0.37 versus 0.32 and 0.31 versus 0.27, respectively).
However, the filter retains only 22\% of
Llama-3.3-70B's in-window facts.
Self-report filtering can therefore leave substantial
historical knowledge in the retained sample or
sharply reduce data coverage.
Its usefulness must be assessed against the measured
actual knowledge boundary.

\input{figures/fig_declared}
\subsection{A declared date changes the words, not the knowledge}
\label{sec:declared}
\label{sec:backtest}


We next test whether declaring a historical date improves the reliability of backtesting by comparing Task~2 \textsc{Predict} with Task~3 in Table~\ref{tab:results}. This comparison shows that declaring a historical date substantially changes models' knowledge claims, while factual accuracy changes little. For GPT-5, the share of answers claiming prior knowledge drops from 0.39 in Task~2 \textsc{Predict} to 0.01 in Task~3, while accuracy remains nearly unchanged (0.76 to 0.75). Similar patterns appear for Llama-3.3-70B (0.63 to 0.00) and GPT-4o (0.33 to 0.04), with no meaningful change in accuracy. DeepSeek-V3.2 shows the opposite trend in knowledge claims (0.37 to 0.46), again with unchanged accuracy. These results show that historical-date instructions change how models describe their answers, not what they can answer.



Table~\ref{tab:backtest} reports Task~3 results inside and beyond each model's measured knowledge window. GPT-5 labels 99\% of its in-window answers as predictions, achieving 0.82 accuracy on these answers, compared with 0.39 after its knowledge boundary. Similar gaps appear for GPT-4o (0.44 versus 0.33) and Llama-3.3-70B (0.51 versus 0.29), with 91\% and 100\% of their in-window answers labeled as predictions, respectively. DeepSeek-V3.2 shows a smaller gap (0.38 versus 0.32), while labeling 46\% of its in-window answers as predictions. These results show that declared-date prompting allows models to present highly accurate answers to previously observed outcomes as forecasts, substantially inflating apparent backtesting performance within their knowledge windows.


\begin{table}[t]
\centering
\small
\setlength{\tabcolsep}{4pt}
\caption{A simulated dated backtest using Task~3 (dated prediction), on controlled facts inside and after each model's knowledge window. \emph{Pred.} is the share of in-window answers labeled as predictions, followed by accuracy on all answers and on those labeled as predictions.}
\label{tab:backtest}
\begin{tabular}{@{}lrrrr@{}}
\toprule
& \multicolumn{3}{c}{\textbf{Inside window}} & \textbf{After}\\
\cmidrule(lr){2-4}\cmidrule(l){5-5}
\textbf{Model} & \textbf{Pred.} & \textbf{Acc (all)} & \textbf{Acc (pred.)} & \textbf{Acc (all)}\\
\midrule
GPT-5         & 0.99 & \textbf{0.83} & \textbf{0.82} & \textbf{0.39}\\
DeepSeek-V3.2 & 0.46 & 0.45 & 0.38 & 0.32\\
GPT-4o        & 0.91 & 0.48 & 0.44 & 0.33\\
Llama-3.3-70B & 1.00 & 0.51 & 0.51 & 0.29\\
\bottomrule
\end{tabular}
\end{table}

\begin{figure}[t]
  \centering
  \includegraphics[width=\columnwidth]{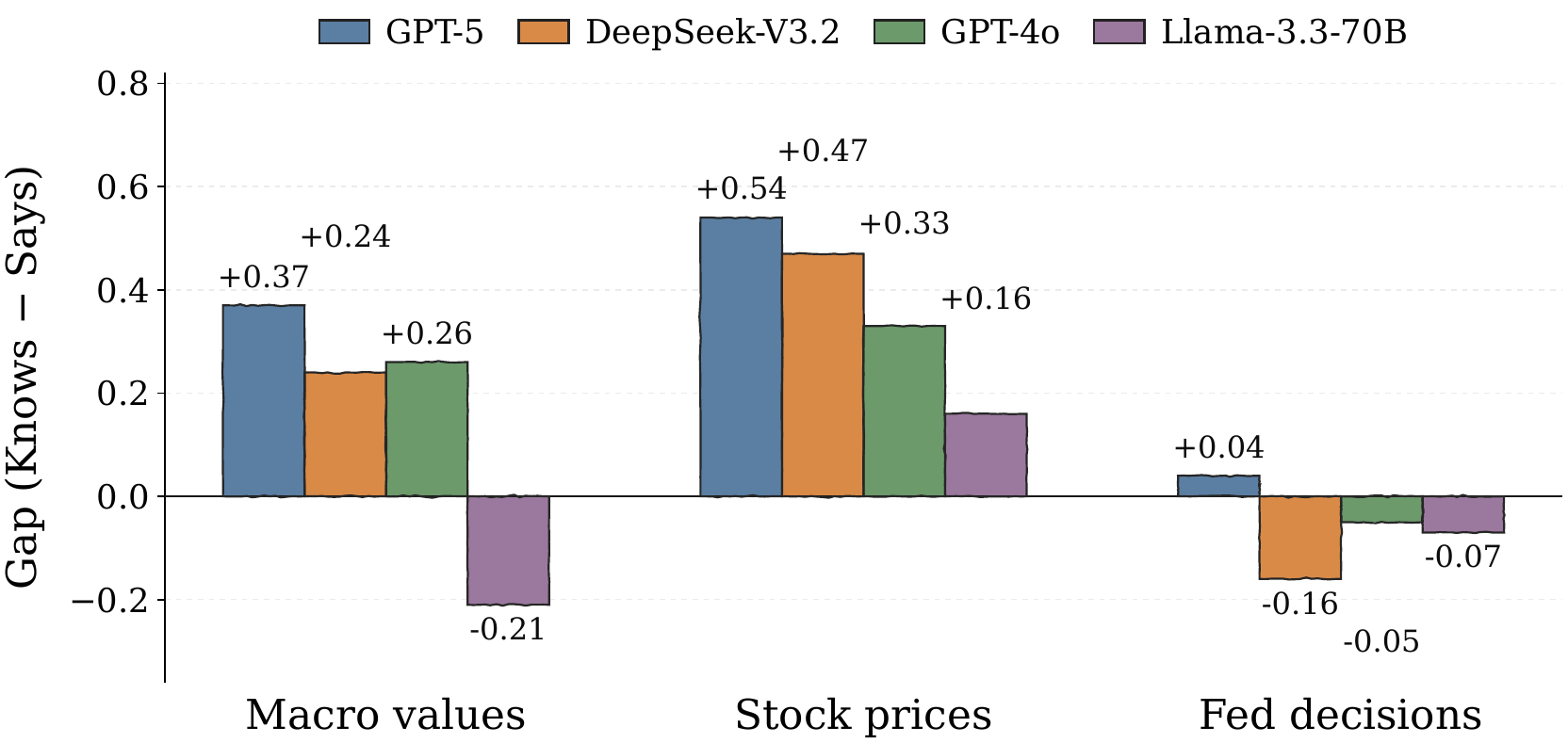}
  \caption{\textbf{Knowledge--claim gaps across financial fact families.}
Each bar shows the difference between Task~1 recall accuracy and the proportion of Task~2 \textsc{Recall} responses claiming knowledge without a declared date.}
  \Description{Three panels, one per fact family, with one row per model. For Fed decisions, knowledge and claims are both near 0.9 for all models. For stock prices, claims are near zero for every model while knowledge is 0.59 for GPT-5, 0.49 for DeepSeek-V3.2 and 0.33 for GPT-4o. For macroeconomic values, claims are below knowledge except for Llama-3.3-70B.}
  \label{fig:family}
\end{figure}


\subsection{Where the hidden knowledge lies}
\label{sec:family}

Figure~\ref{fig:family} shows that knowledge-claim gaps vary substantially across financial fact families. The largest gaps occur for stock prices, where all four models know considerably more than they claim. GPT-5 exhibits a gap of +0.54 (59\% accuracy versus 5\% claimed knowledge), followed by DeepSeek-V3.2 (+0.47; 49\% versus 2\%) and GPT-4o (+0.33; 33\% versus 0\%). Macroeconomic values show a similar pattern for most models, but Llama-3.3-70B exhibits the reverse, claiming knowledge of 53\% of facts while answering only 32\% correctly (-0.21). In contrast, Federal Reserve decisions show much smaller gaps (-0.16 to +0.04), indicating closer agreement between factual accuracy and knowledge claims. These results reveal strong fact-family dependence, with the greatest underclaiming concentrated in precise numerical outcomes.

\subsection{Small and point-in-time models}
\label{sec:small}

Point-in-time models, trained only on data available before fixed cutoffs, offer a potential control for evaluating knowledge boundaries. However, the models we test (1-1.5B parameters) struggle with the multiple-choice format, often selecting the first option regardless of the question. We therefore evaluate TiMaGPT and ChronoGPT across five training cutoffs (2014-2022), along with the 2014 version of DatedGPT, using candidate likelihoods instead. As shown in Table~\ref{tab:pit}, accuracy remains near chance (0.25-0.30) both before and after the training cutoffs. All differences are small ($\leq$0.041), with 95\% bootstrap confidence intervals including zero. These models therefore show no detectable change in factual accuracy around their training boundaries, limiting their usefulness as controls for behavioral cutoff estimation.


\begin{table}[t]
\centering
\small
\setlength{\tabcolsep}{4pt}
\caption{Point-in-time models scored by candidate likelihood. We give Task~1 accuracy on controlled facts up to and after the year of each training cutoff, and the difference with a 95\% bootstrap interval.}
\label{tab:pit}
\begin{tabular}{@{}lrrrl@{}}
\toprule
\textbf{Model} & \textbf{Cutoff} & \textbf{Before} & \textbf{After} & \textbf{Difference}\\
\midrule
TiMaGPT & 2014 & \textbf{0.29} & 0.27 & 0.016 [$-$0.023, 0.055]\\
TiMaGPT & 2016 & 0.28 & 0.26 & 0.021 [$-$0.020, 0.064]\\
TiMaGPT & 2018 & 0.28 & 0.26 & 0.021 [$-$0.020, 0.061]\\
TiMaGPT & 2020 & 0.28 & 0.27 & 0.006 [$-$0.032, 0.048]\\
TiMaGPT & 2022 & 0.27 & 0.26 & 0.006 [$-$0.037, 0.052]\\
ChronoGPT & 2014 & 0.28 & 0.27 & 0.001 [$-$0.038, 0.045]\\
ChronoGPT & 2016 & \textbf{0.29} & 0.27 & 0.023 [$-$0.015, 0.068]\\
ChronoGPT & 2018 & 0.28 & 0.26 & 0.017 [$-$0.019, 0.057]\\
ChronoGPT & 2020 & 0.28 & 0.27 & 0.011 [$-$0.028, 0.051]\\
ChronoGPT & 2022 & 0.27 & \textbf{0.28} & $-$0.010 [$-$0.053, 0.042]\\
DatedGPT & 2014 & \textbf{0.29} & 0.25 & 0.041 [$-$0.003, 0.088]\\
\bottomrule
\end{tabular}
\end{table}

%% file: figures/tab_results.tex
\begin{table}[H]
\centering
\small
\setlength{\tabcolsep}{3.5pt}
\renewcommand{\arraystretch}{1.15}
\caption{Main results by task and wording (8,904 prompts per model). \emph{Knows} is Task~1 accuracy on the 1,340 controlled facts (chance $\approx$ 0.27). In Tasks~2 and~3, \emph{Says} is the share claiming knowledge across the full sample of 2,226 facts, regardless of answer correctness. Task~2 \textsc{Recall} uses no declared date; \emph{Acc$\mid$says} and \emph{Acc$\mid$denies} are Task~1 accuracies on facts grouped by these claims. \emph{Gap} is their within-period accuracy difference, weighted by period size, with 95\% bootstrap intervals. Arrows compare Task~2 \textsc{Predict} (without a declared date) $\rightarrow$ Task~3 (dated prediction), for both \emph{Says} and \emph{Accuracy}. }
\label{tab:results}
\begin{tabular*}{\linewidth}{@{\extracolsep{\fill}}lrrrrrrr@{}}
\toprule
& \multicolumn{1}{c}{\textbf{Task~1}}
& \multicolumn{4}{c}{\textbf{Task~2: \textsc{Recall} claims}}
& \multicolumn{2}{c}{\textbf{Task~2: \textsc{Predict} $\rightarrow$ Task~3}} \\
\cmidrule(lr){2-2}\cmidrule(lr){3-6}\cmidrule(lr){7-8}
\textbf{Model} & \textbf{Knows} & \textbf{Says} & \textbf{Acc$\mid$says} & \textbf{Acc$\mid$denies} & \textbf{Gap [95\% CI]} & \textbf{Says} & \textbf{Accuracy} \\
\midrule
GPT-5                 & \textbf{0.76} & 0.38 & \textbf{0.95} & \textbf{0.61} & 0.18 [0.13, 0.22] & 0.39 $\rightarrow$ 0.01 & \textbf{0.76} $\rightarrow$ \textbf{0.75} \\
DeepSeek-V3.2         & 0.54 & 0.28 & 0.65 & 0.48 & 0.08 [0.03, 0.15] & 0.37 $\rightarrow$ 0.46 & 0.44 $\rightarrow$ 0.43 \\
GPT-4o                & 0.50 & 0.20 & 0.83 & 0.34 & \textbf{0.39} [0.32, 0.45] & 0.33 $\rightarrow$ 0.04 & 0.43 $\rightarrow$ 0.44 \\
Llama-3.3-70B         & 0.43 & 0.48 & 0.58 & 0.29 & 0.28 [0.20, 0.34] & 0.63 $\rightarrow$ 0.00 & 0.41 $\rightarrow$ 0.41 \\
\midrule
Qwen3.5-9B            & 0.37 & 0.69 & 0.40 & 0.30 & 0.10 [0.03, 0.19] & 0.63 $\rightarrow$ \textbf{0.53} & 0.31 $\rightarrow$ 0.35 \\
Qwen3-8B              & 0.36 & \textbf{0.71} & 0.38 & 0.31 & 0.06 [$-$0.01, 0.14] & \textbf{0.68} $\rightarrow$ 0.45 & 0.38 $\rightarrow$ 0.41 \\
\bottomrule
\end{tabular*}
\end{table}


%% file: figures/fig_honesty_map.tex
\begin{figure}[t]
  \centering
  \includegraphics[width=0.7\columnwidth]{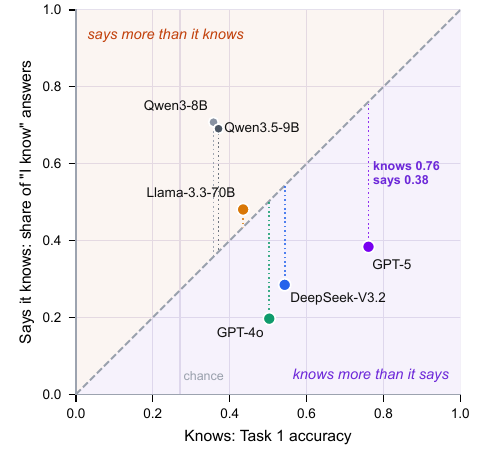}
  \caption{What models know versus what they say. The x-axis is Task~1 accuracy on controlled facts (chance $\approx$ 0.27) and the y-axis is the share of Task~2 \textsc{Recall} answers that claim knowledge. Points below the diagonal under-report their knowledge and points above it over-claim. Dotted lines show each model's distance from the diagonal.}
  \Description{A scatter plot with knowledge on the x-axis and the share of claims on the y-axis. GPT-5, DeepSeek-V3.2, and GPT-4o lie below the diagonal, GPT-5 furthest at 0.76 knowledge and 0.38 claims. Llama-3.3-70B lies just above it at 0.43 and 0.48, and the two Qwen models lie well above it at about 0.36 knowledge and 0.70 claims.}
  \label{fig:honesty}
\end{figure}

%% file: figures/fig_cutoff_timeline.tex
\begin{figure}[t]
  \centering
  \includegraphics[width=\textwidth]{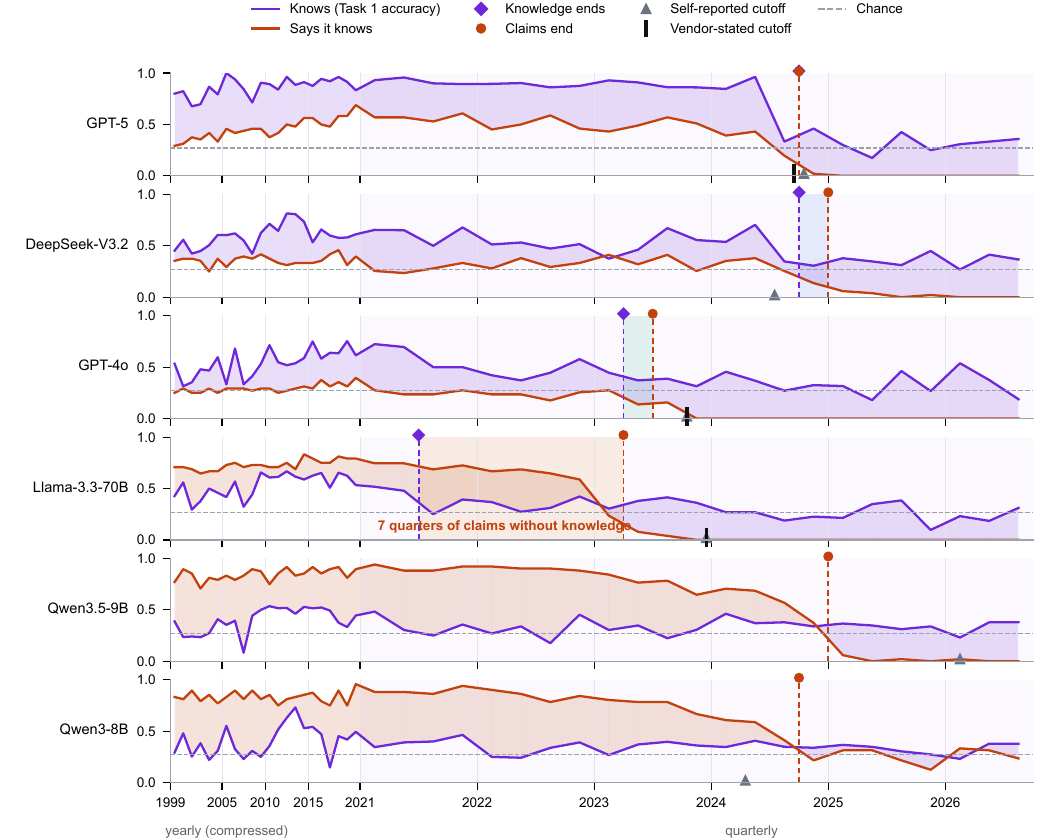}
  \caption{Temporal patterns of factual accuracy and knowledge
claims. Purple and orange lines show Task~1 accuracy and
the share of knowledge claims, respectively. Shading
highlights their differences. Markers indicate the
estimated knowledge boundary ($\blacklozenge$),
claim boundary ($\bullet$), self-reported cutoff
($\blacktriangle$), and vendor-stated cutoff (|),
where available. Years before 2021 are compressed,
while later periods are shown quarterly.}
  \Description{Six stacked time series from 1999 to 2026, one per model. For GPT-5 knowledge and claims drop together in late 2024. For Llama-3.3-70B knowledge drops in mid-2021 while claims stay high until early 2023, leaving a shaded band of about seven quarters. The Qwen models claim knowledge far above their near-chance accuracy until 2024.}
  \label{fig:timeline}
\end{figure}

%% file: sections/08_conclusion.tex
\section{Implications for Backtesting}
\label{sec:implications}

Our findings suggest that financial LLM backtests should incorporate behavioral knowledge audits rather than relying solely on reported cutoffs or temporal prompting. Practitioners can first evaluate models on dated facts relevant to the target domain and estimate their knowledge boundaries using a consistent procedure (Section~\ref{sec:measures}). These estimates can guide the selection of evaluation periods, with appropriate margins for boundary uncertainty. In our experiments, vendor-stated cutoffs are generally conservative relative to measured boundaries. For example, Llama-3.3-70B's reported cutoff excludes ten quarters beyond its measured knowledge boundary, while DeepSeek-V3.2 provides no published cutoff. Self-reported knowledge and historical-date instructions should not replace behavioral measurements, as neither reliably excludes previously known outcomes (Sections~\ref{sec:idk}, and \ref{sec:declared}). Reporting measured boundaries alongside backtesting results, and reassessing them when models are updated, would improve evaluation transparency and comparability.

Our results also have implications for model developers. Knowledge claims are particularly unreliable for precise numerical facts, even when models can identify the correct values (Section~\ref{sec:family}). Moreover, the boundaries of these claims can differ substantially from both demonstrated knowledge and reported cutoffs. Model providers could therefore complement stated training cutoffs with behavioral knowledge assessments on dated, domain-specific facts. Such evaluations would provide practitioners with a clearer picture of the temporal knowledge available to a model and support more reliable financial backtesting.

\section{Conclusion}
\label{sec:conclusion}

We asked whether the safeguards used to keep outcome knowledge out of LLM backtests can be trusted, and built \bench{} to answer from behavior. On dated U.S. financial facts, the answer is no.
Frontier models know more than they say, stop claiming knowledge at a date they believe rather than at the end of what they retain, answer many of the facts they disclaim, and drop their claims of prior knowledge when told to answer as of a past date without forgetting anything. 
A stated cutoff, a self-report and a declared date each fail to separate known from unknown outcomes. Backtests should therefore be restricted to periods after a knowledge boundary measured on the model itself, or run on models whose training data end before the test period. \bench{} provides the facts, questions and procedure for the first, and shows that current point-in-time models are still too small for the second.

%% file: sections/07_analysis.tex
\section*{Limitations and Ethics}

\paragraph{Limitations} \bench{} covers one economy and three families of monthly financial facts, and boundaries for other regions, other kinds of financial knowledge or text-based events may differ. The boundary rule uses fixed thresholds relative to each model's 2015 to 2020 level and chance. Other thresholds move the estimated quarters, although the same rule is applied to every model. A model can be right without recall when it extrapolates a trend, and our persistence control removes only the simplest such strategy. Frontier models were accessed through an API router and score one response per prompt at temperature~0 where supported. Served versions may change over time, and we excluded one model whose served identity we could not verify. Free-form or probabilistic statements of knowledge may behave differently from our multiple-choice claims. Finally, the point-in-time models we test are small, so they cannot yet show whether a large model with a strict cutoff would behave honestly.

\paragraph{Ethics} All facts come from public sources (FRED, published exchange quotes and Federal Reserve announcements) and contain no personal data. The benchmark is meant to make financial evaluations of LLMs more reliable, and its main risk is misuse as a certificate that a model is leak-free for a period, which our results caution against.

%% file: sections/A_data_construction.tex
\section{Data Collection and Construction}
\label{app:data-construction}

\subsection{Sources and Historical Values}

The fact pool contains 20,972 dated U.S. financial
facts from January 1999 to September 2026, covering
macroeconomic indicators, company stock prices,
and Federal Reserve policy outcomes.

\paragraph{Macroeconomic facts.}
We collect monthly observations for twelve FRED
series: headline and core CPI inflation,
unemployment, the effective federal funds rate,
the 3-month, 2-year, and 10-year Treasury rates,
the 30-year mortgage rate, three dollar exchange
rates, and WTI oil.
First-release observations are retrieved through
ALFRED, retaining the initial publication date
and value rather than replacing them with
subsequent revisions.\footnote{
\url{https://fred.stlouisfed.org/};
\url{https://alfred.stlouisfed.org/}.}
The question specifies the indicator and reference
period, and candidate values use the corresponding
units and numerical precision.
For example, the illustrative CPI fact in
Table~\ref{tab:example} concerns the 12-month change
in CPIAUCNS for June 2023, rounded to one decimal
place.

\paragraph{Company facts.}
We collect month-end closing prices for 50 large
U.S. companies from Yahoo Finance using
\texttt{yfinance}.\footnote{
\url{https://finance.yahoo.com/};
\url{https://ranaroussi.github.io/yfinance/}.}
For each reference month, the observation is
the closing price on its last trading day.
Subsequent stock-split adjustments are reversed
to recover the price as quoted at the time.

\paragraph{Policy facts.}
We derive monthly Federal Reserve target-rate
outcomes from the target-rate history available
through FRED.
Each fact has one of three outcomes:
raised, lowered, or unchanged.
The source series cover both the earlier
single target rate and the later target
range.\footnote{
\url{https://fred.stlouisfed.org/series/DFEDTAR};
\url{https://fred.stlouisfed.org/series/DFEDTARL};
\url{https://fred.stlouisfed.org/series/DFEDTARU}.}



\subsection{Sampling}

The evaluation sample contains 2,226 facts:
1,353 macroeconomic observations, 540 company
stock prices, and 333 policy outcomes.
Sampling is sparse before 2021, using two
macroeconomic observations and one stock price
per month, together with the monthly policy
outcomes.
From 2021 onward, sampling covers every
macroeconomic series and four stock prices
per month, together with the policy outcomes.
This allocates more observations to the period
containing the cutoffs of the evaluated models.
The resulting sample contains 1,056 facts
before 2021 and 1,170 from 2021 onward.
Table~\ref{tab:stats} gives the counts by family. 
It should be noted that the October 2025 observations for headline inflation,
core inflation, and unemployment are unavailable
because of the U.S. federal government shutdown.

\subsection{Candidate Values and Prompt Construction}

For each numerical fact, we construct four
candidate values with spacing approximately
equal to one typical monthly change of the
corresponding series.
The rank of the correct value among the
ordered candidates is drawn uniformly.
Policy facts instead use the three outcomes
raised, lowered, and unchanged.

Each fact produces four prompts across
three tasks.
Task~1 presents the candidate values directly.
Task~2 presents the same candidates under
\textsc{Recall} and \textsc{Predict} wordings.
In both wordings, each candidate appears with
two knowledge claims: knowing the value and
not knowing it.
This produces eight options for numerical
facts and six for policy facts.
Task~3 retains the Task~2 \textsc{Predict}
question and paired options, adding the
historical-date instruction described above.

The underlying fact and candidate values
are shared across all four prompts.
Each prompt is sent independently with
shuffled options and requests a single
option letter.
The resulting evaluation comprises
$2{,}226 \times 4 = 8{,}904$ prompts per model.


\subsection{Controlled Evaluation Subset}

The full sample is retained for prompt
construction and knowledge-claim measurement.
For the factual knowledge score, we apply
the controls in Section~\ref{sec:controls}.
Numerical evaluation retains 1,007 facts
after the persistence control, comprising
715 macroeconomic observations and
292 stock prices.
Together with all 333 policy facts, these
form the 1,340-fact controlled subset.
Policy performance is measured using
balanced accuracy across the three outcomes.
Thus, the controls determine how factual
knowledge is measured without changing the
four-prompt construction for each fact.

%% file: sections/B_cutoff_summary.tex
\section{Cutoff Summary}
\label{app:cutoff-summary}

Table~\ref{tab:cutoffs} collects the estimated
boundaries and reported cutoffs for the four
frontier models.
Following Section~\ref{sec:measures},
\emph{knowledge ends} is estimated from
controlled Task~1 accuracy, whereas
\emph{claims end} is estimated from
Task~2 \textsc{Recall} knowledge claims.

Let $B_K$ denote a model's 2015--2020 Task~1
accuracy under the controls in
Section~\ref{sec:controls}, and let $C \approx 0.27$
denote the combined chance reference.
The default knowledge threshold is
\[
T_K = C + \frac{1}{2}\bigl(B_K - C\bigr).
\]
Following Section~\ref{sec:measures},
\emph{knowledge ends} is the last quarter whose
two-quarter rolling accuracy (\emph{Knows})
stays above $T_K$.

\begin{table}[t]
\centering
\small
\setlength{\tabcolsep}{7pt}
\caption{Estimated boundaries and reported cutoffs
for the four frontier models.
\emph{Knowledge ends} and \emph{Claims end}
follow the estimation procedure in
Section~\ref{sec:measures}.
\emph{Self-reported} is the cutoff given when
the model is asked directly;
\emph{Vendor-stated} is the date published
by its developer.
``n/a'' indicates that no vendor-stated
date is recorded.}
\label{tab:cutoffs}
\begin{tabular}{@{}lllll@{}}
\toprule
\textbf{Model} & \textbf{Knowledge ends} & \textbf{Claims end}
& \textbf{Self-reported} & \textbf{Vendor-stated} \\
\midrule
GPT-5
& 2024Q3 & 2024Q3 & 2024-10 & 2024-09 \\
DeepSeek-V3.2
& 2024Q3 & 2024Q4 & 2024-07 & n/a \\
GPT-4o
& 2023Q1 & 2023Q2 & 2023-10 & 2023-10 \\
Llama-3.3-70B
& 2021Q2 & 2023Q1 & 2023-12 & 2023-12 \\
\bottomrule
\end{tabular}
\end{table}

%% file: sections/C_manual_validation.tex
\section{Manual Data Validation}
\label{app:manual-validation}

\paragraph{Review setup.}
We conducted a manual review to examine the reliability
of the historical facts and the quality of the questions
constructed from them. Two reviewers examined the same
set of 50 facts using the four criteria in
Table~\ref{tab:manual-checks}.
Factual correctness requires source support for the
entity, indicator, reference period, units, calculations,
and rounding. First-release and temporal validity require
historical evidence for the initial published value and
distinguish the reference period, first publication date,
and hypothetical date in Task~3. Question clarity covers
the specification of the requested quantity and its
measurement conventions. Option validity and task
consistency cover a unique correct candidate, distinct
distractors, correct answer mappings, and shared facts
and candidate values across the four prompts.

\paragraph{Review results.}
Both reviewers judged all 50 reviewed facts to satisfy
the four criteria in Table~\ref{tab:manual-checks},
yielding a final pass rate of 100\% within the reviewed
sample. The review identified no issues in the factual
values or temporal references, and no concerns about
the clarity of the questions or the validity and task
consistency of the answer options.

\begin{table}[t]
\centering
\small
\setlength{\tabcolsep}{5pt}
\renewcommand{\arraystretch}{1.15}
\caption{Manual validation criteria.
An item passes a dimension when it meets
all criteria listed in that row.}
\label{tab:manual-checks}
\begin{tabular}{@{}
>{\raggedright\arraybackslash}p{0.25\textwidth}
>{\raggedright\arraybackslash}p{0.70\textwidth}
@{}}
\toprule
Dimension & Passing criteria \\
\midrule

Factual correctness
&
A reliable source supports the fact.
The entity, indicator, reference period,
units, calculations, and rounding are correct.
\\

First-release and temporal validity
&
Historical evidence supports the gold answer
as the first-release value.
The reference period, first publication date,
and hypothetical date in Task~3 are correctly
identified and distinguished.
\\

Question clarity
&
The question uniquely identifies the intended
fact.
Relevant conventions are explicit, including
year-over-year versus month-over-month changes,
monthly averages versus month-end values,
seasonal adjustment, and stock-price adjustments.
\\

Option validity and task consistency
&
The correct factual candidate is present and
unique; distractors are distinct, and gold
option mappings are correct.
The four prompts refer to the same fact and
share the same candidate values.
Each candidate has both knowledge-claim
options in Tasks~2 and~3.
\\

\bottomrule
\end{tabular}
\end{table}

%% file: sections/D_model_details.tex
\section{Evaluated Models}
\label{sec:evaluated-models}

\begin{table}[!t]
\centering
\scriptsize
\setlength{\tabcolsep}{3pt}
\renewcommand{\arraystretch}{1.0}

\caption{Models evaluated in \bench{}. Parameter counts are
reported in billions (B), with active parameters given
for mixture-of-experts models where available.
Cutoff refers to the vendor-reported knowledge cutoff
for frontier models and the training-data cutoff for
point-in-time models. NR indicates information not
reported or verified.}

\label{tab:model_list}

\begin{tabularx}{\textwidth}{
@{}lclcl>{\raggedright\arraybackslash}Xl@{}
}
\toprule
Model & Params. & Architecture & Cutoff &
Access & HF Repository ID & License \\
\midrule

\multicolumn{7}{@{}l}{\textbf{Frontier models}}\\
\addlinespace[2pt]

GPT-5
& NR & Proprietary & 2024-09
& API & --- & Proprietary \\

DeepSeek-V3.2
& 685B (37B active) & MoE & NR
& API
& \href{https://huggingface.co/deepseek-ai/DeepSeek-V3.2}
  {\nolinkurl{deepseek-ai/DeepSeek-V3.2}}
& MIT \\

GPT-4o
& NR & Proprietary & 2023-10
& API & --- & Proprietary \\

Llama-3.3-70B-Instruct
& 70B & Dense & 2023-12
& API
& \href{https://huggingface.co/meta-llama/Llama-3.3-70B-Instruct}
  {\nolinkurl{meta-llama/Llama-3.3-70B-Instruct}}
& Llama 3.3 \\

\midrule

\multicolumn{7}{@{}l}{\textbf{Open models}}\\
\addlinespace[2pt]

Qwen3-8B
& 8B & Dense & NR
& Local
& \href{https://huggingface.co/Qwen/Qwen3-8B}
  {\nolinkurl{Qwen/Qwen3-8B}}
& Apache 2.0 \\

Qwen3.5-9B
& 9B & Hybrid & NR
& Local
& \href{https://huggingface.co/Qwen/Qwen3.5-9B}
  {\nolinkurl{Qwen/Qwen3.5-9B}}
& Apache 2.0 \\

\midrule

\multicolumn{7}{@{}l}{\textbf{Point-in-time models}}\\
\addlinespace[2pt]

TiMaGPT-2014
& NR & GPT-2 & 2014
& Local
& \href{https://huggingface.co/Ti-Ma/TiMaGPT2-2014}
  {\nolinkurl{Ti-Ma/TiMaGPT2-2014}}
& CC0 1.0 \\

TiMaGPT-2016
& NR & GPT-2 & 2016
& Local
& \href{https://huggingface.co/Ti-Ma/TiMaGPT2-2016}
  {\nolinkurl{Ti-Ma/TiMaGPT2-2016}}
& CC0 1.0 \\

TiMaGPT-2018
& NR & GPT-2 & 2018
& Local
& \href{https://huggingface.co/Ti-Ma/TiMaGPT2-2018}
  {\nolinkurl{Ti-Ma/TiMaGPT2-2018}}
& CC0 1.0 \\

TiMaGPT-2020
& NR & GPT-2 & 2020
& Local
& \href{https://huggingface.co/Ti-Ma/TiMaGPT2-2020}
  {\nolinkurl{Ti-Ma/TiMaGPT2-2020}}
& CC0 1.0 \\

TiMaGPT-2022
& NR & GPT-2 & 2022
& Local
& \href{https://huggingface.co/Ti-Ma/TiMaGPT2-2022}
  {\nolinkurl{Ti-Ma/TiMaGPT2-2022}}
& CC0 1.0 \\

ChronoGPT-Instruct-2014
& 1.55B & Dense & 2014
& Local
& \href{https://huggingface.co/manelalab/chrono-gpt-instruct-v1-20141231}
  {\nolinkurl{manelalab/chrono-gpt-instruct-v1-20141231}}
& MIT \\

ChronoGPT-Instruct-2016
& 1.55B & Dense & 2016
& Local
& \href{https://huggingface.co/manelalab/chrono-gpt-instruct-v1-20161231}
  {\nolinkurl{manelalab/chrono-gpt-instruct-v1-20161231}}
& MIT \\

ChronoGPT-Instruct-2018
& 1.55B & Dense & 2018
& Local
& \href{https://huggingface.co/manelalab/chrono-gpt-instruct-v1-20181231}
  {\nolinkurl{manelalab/chrono-gpt-instruct-v1-20181231}}
& MIT \\

ChronoGPT-Instruct-2020
& 1.55B & Dense & 2020
& Local
& \href{https://huggingface.co/manelalab/chrono-gpt-instruct-v1-20201231}
  {\nolinkurl{manelalab/chrono-gpt-instruct-v1-20201231}}
& MIT \\

ChronoGPT-Instruct-2022
& 1.55B & Dense & 2022
& Local
& \href{https://huggingface.co/manelalab/chrono-gpt-instruct-v1-20221231}
  {\nolinkurl{manelalab/chrono-gpt-instruct-v1-20221231}}
& MIT \\

DatedGPT-Instruct-2014
& 1.3B & Dense & 2014
& Local
& \href{https://huggingface.co/datedgpt/datedgpt-2014-instruct}
  {\nolinkurl{datedgpt/datedgpt-2014-instruct}}
& NR \\

\bottomrule
\end{tabularx}
\end{table}

Table~\ref{tab:model_list} lists the 17 model variants we evaluated. The main experiments cover four frontier models: GPT-5~\cite{openai2025gpt5}, DeepSeek-V3.2~\cite{deepseek2025v32}, GPT-4o~\cite{openai2024gpt4o}, and Llama-3.3-70B-Instruct~\cite{meta2024llama33}. We also include two smaller open models, Qwen3-8B~\cite{yang2025qwen3} and Qwen3.5-9B~\cite{qwen2026qwen35}, to compare their factual knowledge and self-reports with those of the frontier models. The four frontier models are accessed through APIs, while the Qwen models are run locally. Model names and reported specifications follow the developers' documentation, and the table links to the corresponding Hugging Face repositories where weights are available.

We further evaluate three families of point-in-time models: TiMaGPT~\cite{drinkall2024timemachine}, ChronoGPT-Instruct~\cite{he2025chronologically}, and DatedGPT~\cite{yan2026datedgpt}. Unlike general-purpose models, these models are trained on text restricted to specific historical periods, making them useful comparisons for studying temporal knowledge boundaries. We use five checkpoints each for TiMaGPT and ChronoGPT-Instruct, with cutoffs spanning 2014 to 2022, and the 2014 checkpoint of DatedGPT. Because these smaller models often fail to follow the multiple-choice format, we score candidate likelihoods instead of generated answers (Section~\ref{sec:small}). The table also records the licenses stated in the model repositories, with unavailable information marked as NR.

\input{figures/model_list}